%% file: iclr2023_conference.tex
\documentclass{article} 
\usepackage{iclr2023_conference,times}

\input{math_commands.tex}

\usepackage{hyperref}
\usepackage{url}
\usepackage{booktabs}       
\usepackage{amsfonts}       
\usepackage{nicefrac}       
\usepackage{microtype}      
\usepackage{xcolor}         
\usepackage{amsmath}
\usepackage{amssymb}
\usepackage{mathtools}
\usepackage{amsthm}
\usepackage{multirow}
\usepackage{wrapfig,lipsum,booktabs}
\usepackage{caption}
\usepackage{amsthm,amsmath,amssymb}
\usepackage{mathrsfs}
\usepackage{xcolor}
\usepackage{dsfont}
\usepackage{diagbox}

\title{Accurate and Reliable Predictions with Mutual-Transport Ensemble}

\author{
  Han Liu$^{\dagger}$\\
  Tsinghua University\\
  \texttt{han-liu18@mails.tsinghua.edu.cn} \\
  \And
  Peng Cui$^{\dagger}$\\
  Tsinghua University\\
  \texttt{xpeng.cui@gmail.com} \\
  \And
  Bingning Wang \\
  daniel@baichuan-inc \\
  \texttt{daniel@baichuan-inc.com} \\
  \And
  Jun Zhu\\
  Tsinghua University\\
  \texttt{dcszj@tsinghua.edu.cn} \\
  \And
  Xiaolin Hu$^{*}$ \\
  Tsinghua University \\
  \texttt{xlhu@tsinghua.edu.cn} \\
}
\footnotetext{$\dagger$ These authors contributed equally to this work.}
\footnotetext{* Corresponding author}
\iclrfinalcopy
\begin{document}

\maketitle

\begin{abstract}
Deep Neural Networks (DNNs) have achieved remarkable success in a variety of tasks, especially when it comes to prediction accuracy. However, in complex real-world scenarios, particularly in safety-critical applications, high accuracy alone is not enough. Reliable uncertainty estimates are crucial. Modern DNNs, often trained with cross-entropy loss, tend to be overconfident, especially with ambiguous samples.
To improve uncertainty calibration, many techniques have been developed, but they often compromise prediction accuracy. To tackle this challenge, we propose the ``mutual-transport ensemble'' (MTE). This approach introduces a co-trained auxiliary model and adaptively regularizes the cross-entropy loss using Kullback-Leibler (KL) divergence between the prediction distributions of the primary and auxiliary models.
We conducted extensive studies on various benchmarks to validate the effectiveness of our method. The results show that MTE can simultaneously enhance both accuracy and uncertainty calibration. For example, on the CIFAR-100 dataset, our MTE method on ResNet34/50 achieved significant improvements compared to previous state-of-the-art method, with absolute accuracy increases of $2.4\%/3.7\%$, relative reductions in ECE of $42.3\%/29.4\%$, and relative reductions in classwise-ECE of $11.6\%/15.3\%$.
\end{abstract}

\section{Introduction}
\label{sec:intro}
Deep Neural Networks (DNNs) have achieved great success in various tasks, especially in terms of prediction accuracy. In safety-critical areas such as medical diagnosis~\citet{leibig2017leveraging,dolezal2022uncertainty}, self-driving cars~\citet{michelmore2020uncertainty}, and protein engineering~\citet{greenman2023benchmarking}, accurate uncertainty estimation (i.e., well-calibrated prediction) is crucial for reliable model deployment. A reliable model needs to show high uncertainty when making incorrect predictions.

The mainstream method for training classification models is cross-entropy (CE) loss, which has been widely adopted due to its effectiveness in achieving high accuracy.
Nonetheless, DNNs trained with CE loss tend to become overconfident, especially when dealing with ambiguous samples, leading to predictions that are often close to 100\% confident but unreliable~\citet{guo2017calibration,cui2024learning}.

Some studies try to solve this with various regularization methods during training, known as train-time calibration.
For example, Entropy Regularization (ER)~\citet{pereyra2017regularizing} and Label Smoothing (LS)~\citet{muller2019does} encourage smoother predicted probabilities to reduce overconfidence with maximized entropy or smoothed labels. MMCE~\citet{mukhoti2020calibrating} minimizes calibration error with a measure from kernel mean embeddings, which keeps the overconfident predictions. To improve the confidence of predicted labels, MDCA~\citet{hebbalaguppe2022stitch} uses a secondary loss function to calibrate the whole predicted probability distribution. While these methods reduce overconfidence, they often hurt prediction accuracy due to strong regularization.

Deep Ensembles (DE)~\citet{lakshminarayanan2017simple,fort2019deep} offers a solution by combining multiple randomly initialized trained DNNs to improve both prediction accuracy and uncertainty calibration. But using multiple DNNs during training and inference is computationally expensive, limiting real-world use.

To address these challenges, we introduce a method named ``\textit{Mutual-Transport Ensemble}'' (MTE), which draws inspiration from DE but combines elements of regularization and ensemble techniques in a unique way. MTE introduces a co-trained auxiliary model that utilizes the prediction distribution of the primary model as supervised labels. The Kullback-Leibler (KL) divergence between the prediction distributions of the primary and auxiliary models serves as an adaptive regularizer, preserving diversity and mitigating overfitting to one-hot ground-truth labels.
One advantage of MTE over DE is its reduced computational overhead during the inference phase. 

\begin{figure*}[t]
  \centering
  \includegraphics[width=\textwidth]{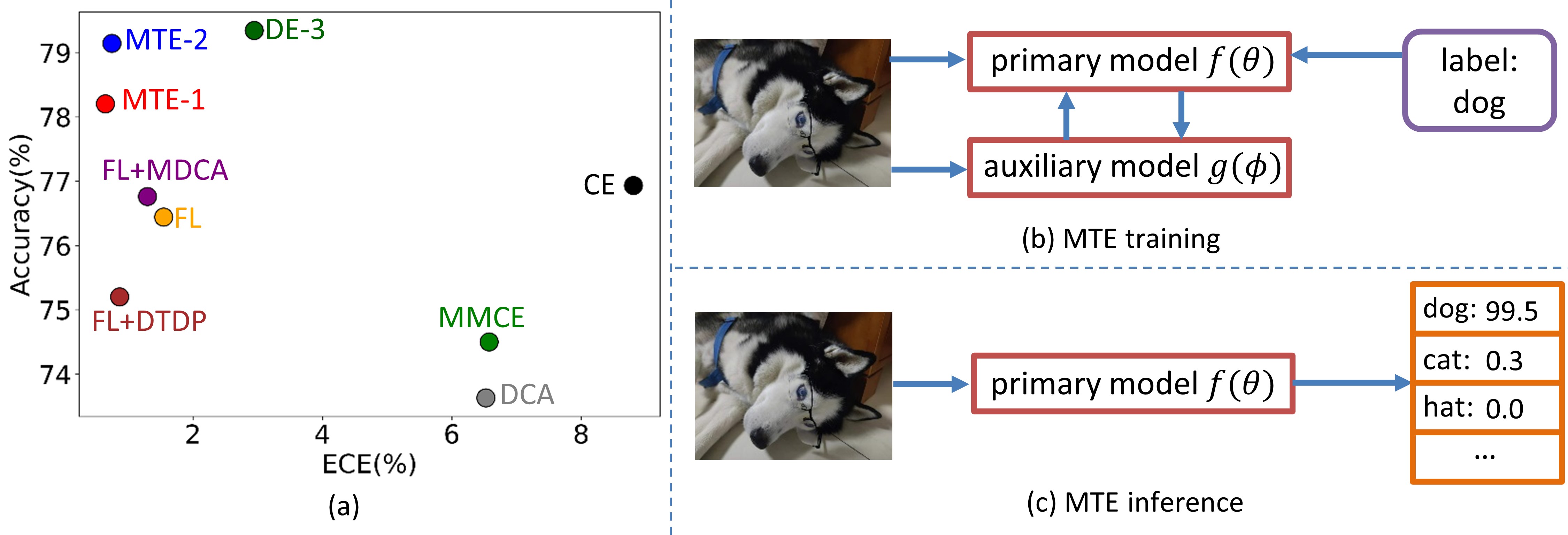}
  \caption{\small{(a) Scatter plot of ECE and accuracy for different calibration methods based on the ResNet34 backbone on the CIFAR-100 test set. Methods located closer to the top left corner perform better. Our MTE method is positioned at the top left corner, indicating the best performance.
(b) Schematic diagram of the training process for MTE.
(c) Schematic diagram of the inference process for MTE.}}
\label{fig:methods}
\end{figure*}

We tested our method extensively on various benchmarks, including image classification with and without distributional shift, misclassification detection, and out-of-distribution (OOD) detection. Unlike previous methods that lose accuracy or require heavy computation, our method improves both prediction performance and uncertainty estimation without being computationally intensive during inference. 
For example, on the CIFAR-100 dataset, our MTE method on ResNet34/50 achieved significant improvements compared to the previous state-of-the-art method, MDCA~\citet{hebbalaguppe2022stitch}, with absolute accuracy increases of $2.4\%/3.7\%$, relative reductions in ECE of $42.3\%/29.4\%$, and relative reductions in classwise-ECE of $11.6\%/15.3\%$. These results clearly demonstrate the effectiveness of our method in enhancing both prediction accuracy and uncertainty calibration.
Figure~\ref{fig:methods}(a) presents the scatter plot of ECE-accuracy for different calibration methods.
Additionally, we conducted thorough ablation studies to check the robustness of our method with different hyper-parameters. These studies show the stability and reliability of our approach in various settings.

\section{Background and Related Work}
Calibration, the process of ensuring that the predicted probabilities reflect the true likelihood of the events, plays a crucial role in machine learning models. It ensures that the prediction of machine learning models are not only accurate but also well-calibrated. Well-calibrated models provide reliable uncertainty estimates, which can help users and downstream systems make safer decisions by avoiding overconfidence.

There are two fundamental types of calibration: confidence calibration~\citet{naeini2015obtaining} and classwise calibration~\citet{kull2019beyond}.
In this paper, we focus on both types of calibration.

\textbf{Confidence Calibration}:
Confidence calibration refers to the alignment of predicted confidence with observed frequencies. It ensures that a model's prediction confidence accurately reflect the true likelihood of an event occurring. 
A widely used metric for evaluating confidence calibration is the Expected Calibration Error (ECE)~\citet{naeini2015obtaining}. 
The definition of ECE can be found in Appendix~\ref{sec:background appendix}.

\textbf{Classwise Calibration}:
Classwise calibration is concerned with calibrating the predicted probabilities for each individual class in a multi-class classification problem. It ensures that the predicted probabilities across different classes are accurate and properly calibrated. 
The metric for classwise calibration is classwise-ECE~\citet{kull2019beyond}, which can be found in Appendix~\ref{sec:background appendix}.

According to the number of models used for inference, calibration methods can be divided into single-model calibration methods and Deep Ensembles (DE)~\citet{lakshminarayanan2017simple,fort2019deep}. 

Within single-model calibration methods, there are two categories: train-time calibration methods and post-hoc calibration methods. Train-time calibration methods typically involve designing loss functions or incorporating additional regularization techniques, such as
Entropy Regularization (ER)~\citet{pereyra2017regularizing}, Label Smoothing (LS)~\citet{muller2019does,szegedy2016rethinking}, Focal Loss (FL)~\citet{mukhoti2020calibrating}, Maximum Mean Calibration Error (MMCE)~\citet{kumar2018trainable}, Difference between Confidence and Accuracy (DCA)~\citet{liang2020improved}, Multi-class Difference in Confidence and Accuracy (MDCA)~\citet{hebbalaguppe2022stitch}, and Dynamic Train-time Data Pruning (DTDP)~\citet{patra2023calibrating}.
Post-hoc calibration methods are applied after the model has been trained, such as Temperature Scaling (TS)~\citet{platt1999probabilistic}, Dirichlet calibration (DC)~\citet{kull2019beyond}, and Meta-calibration~\citet{bohdal2021meta}. 

DE creates an ensemble by training multiple models independently with different initializations. During inference, the predictions from all models are combined, typically by averaging their predicted probabilities. While effective for enhancing model performance, DE introduces additional computational overhead during inference.

For a detailed introduction to related works, please refer to Appendix~\ref{sec:related work}.

\section{The MTE Method}
\label{sec:method}
Let $P_{data}$ denote the probability distribution of data. A dataset consists of finite $(x,y_x)$ pairs where $x\sim P_{data}$ is a data instance and $y_x \in \mathcal{Y}=\{1,2,...,K\}$ is the ground-truth class label of $x$.
A model takes in the instance $x$ and output its prediction vector $p_x\in \mathbb{R}^{K}$. Let $p^{(k)}_x$ denote the $k$-th dimension of $p_x$.
The confidence $\hat{p}_x=\text{max}_{k\in \mathcal{Y}}\{p^{(k)}_x\}$ is the maximum value in the probability vector of predictions, and the model's prediction $\hat{y}_x=\text{argmax}_{k\in \mathcal{Y}}\{p^{(k)}_x\}$ is the index corresponding to the maximum value.

Typically, we train a model by minimizing the following cross-entropy (CE) loss on the training set,
\begin{equation}
\begin{aligned}
L_{\text{CE}}(p_x)=-\log p^{(y_x)}_x,
\label{equ:ce}
\end{aligned}
\end{equation}
where $p^{(y_x)}_x$ represents the prediction probability corresponding to the ground-truth label $y_x$. Previous studies~\citet{guo2017calibration,mukhoti2020calibrating,cui2024learning} reveal that minimizing the CE loss forces the model to overfit to one-hot encoded ground-truth labels with almost 100\% probability irrespective of their correctness, which can result in overconfidence. 

To address this issue, we propose a dual-model training method MTE, which consists of a primary model $f(\theta)$ and an auxiliary model $g(\phi)$. Here, $\theta$ and $\phi$ denote the parameters of $f$ and $g$ respectively. In MTE, the auxiliary model leverages its predicted probability distribution to adaptively coach the primary model in producing the well-calibrated prediction. The schematic diagrams of MTE training and inference are illustrated in Figure~\ref{fig:methods}(b) and (c), respectively. 
\label{sec:method}

\subsection{Mutual-Transport Ensemble}
In order to guide the training of the primary model, we let the primary model interact with the auxiliary model by incorporating the KL divergence of predicted probability distributions of two models into the training objective. We denote the predicted probability distributions of the primary model and the auxiliary model given an input $x$ as $f_x(\theta)$ and $g_x(\phi)$, respectively.
Kullback-Leibler (KL) divergence is a measure of how one probability distribution diverges from another probability distribution:
\begin{equation}
\begin{aligned}
D_{\text{KL}}(g_x(\phi)||f_x(\theta))=\sum_{k=1}^K g_x^{(k)}(\phi)\log \frac{g_x^{(k)}(\phi)}{f_x^{(k)}(\theta)}.
\label{equ:KL}
\end{aligned}
\end{equation}

\textbf{Train Primary Model $f(\theta)$}: We propose to regularize the cross-entropy loss with the discrepancy between the predictions of the primary model and the auxiliary model. We can optimize $f(\theta)$ by the following loss function:
\begin{equation}
\begin{aligned}
L_{f,x}(\theta)=L_{\text{CE}}(f_x(\theta))+\alpha D_{\text{KL}}(g_x(\phi)||f_x(\theta)).
\label{equ:loss A}
\end{aligned}
\end{equation}
The hyper-parameter $\alpha$ is a balancing factor.
The first term in loss~\eqref{equ:loss A} is equivalent to $D_{KL}(T_x || f_x(\theta))$ where $T_x$ is the ground truth label (a one-hot vector) for input x.
In other words, the first term forces $f_x(\theta)$ to approach an extremely stiff target while the second term forces it to approach a smoother target $g_x(\phi)$, an output of softmax function. In this way we prevent $f(\theta)$ from overfitting a one-hot ground truth label and encourage it to learn a well-calibrated probability distribution. 

\textbf{Train Auxiliary Model $g(\phi)$}: We hope that the auxiliary model is not misled by underlying ambiguous ground truth existing in training data and can provide reliable supervised information for the primary model as much as possible. To this end, we utilize the output probability of the primary model as soft labels to train the auxiliary model. Concretely, we minimize the KL divergence between two probability distributions predicted by the model $f_x(\theta)$ and $g_x(\phi)$ to optimize $g(\phi)$:
\begin{equation}
\begin{aligned}
L_{g,x}(\phi)=D_{\text{KL}}(f_x(\theta)||g_x(\phi)).
\label{equ:loss B}
\end{aligned}
\end{equation}
During the training stage, the primary model and the auxiliary model are trained simultaneously, learning from each other.
Inherently, through the KL divergence term in loss function \eqref{equ:loss A} and \eqref{equ:loss B}, we construct a mutual feedback loop between the primary and auxiliary models. The collaborative learning and mutual feedback not only boost the prediction diversity of the primary model but also assist the auxiliary model in learning a dependable prediction distribution to regularize the cross-entropy loss.

During the inference stage, a single primary model suffices for making predictions in our method, unlike DE, which requires combining outputs of multiple models simultaneously. Hence, the proposed method significantly reduces computational overhead.

\subsection{Extending to Multiple Auxiliary Models}
Naturally, we can incorporate multiple auxiliary models $\{g_i\}_{i=1}^{N_g}$ into the training process to further improve the prediction performance and calibration of the primary model $f$. Analogous to DE, leveraging multiple auxiliary models can effectively promote model diversity and improve the uncertainty calibration of the primary model. The optimization objectives for the primary model and each $g_i$ are formalized as:
\begin{equation}
\begin{aligned}
L_{f,x}(\theta)=L_{\text{CE}}(f_x(\theta))+\alpha \sum_{i=1}^{N_g}\frac{1}{N_g}D_{\text{KL}}(g_{i,x}(\phi_i)||f_x(\theta)),
\label{equ:loss A multi}
\end{aligned}
\end{equation}
\begin{equation}
\begin{aligned}
L_{g_i,x}(\phi_i)=D_{\text{KL}}(f_x(\theta)||g_{i,x}(\phi_i)),
\label{equ:loss B multi}
\end{aligned}
\end{equation}
where $\phi_i$ denotes the trainable parameters of the auxiliary model $g_i$.

\subsection{The Connection between MTE and DE}
DE, a pioneering work, has demonstrated superior performance in both accuracy and calibration by exploiting the diversity among independently trained models. Building on this concept, MTE further enhances these aspects by incorporating mutual information and adaptive regularization. To better understand the principles behind MTE, we analyze our method from a theoretical perspective, highlighting how it extends the strengths of DE to achieve improved prediction accuracy and uncertainty calibration.
For DE, given two trained models $f$ and $g$ and an input $x$, we can get a weighted sum of predictive probabilities from $f_x(\theta)$ and $g_x(\phi)$ as the final prediction:
\begin{equation}
\begin{aligned}
h_x(\theta,\phi)=(1-\beta)f_x(\theta)+\beta g_x(\phi),
\label{equ:ensemble p}
\end{aligned}
\end{equation} 
where $\beta \in [0,1]$ is a weight to balance the effect of each model.

To explain the relationship between MTE and DE, we first present a proposition (proof is in Appendix~\ref{sec:proof}).

\textbf{Proposition 1}. \textit{Given two prediction probability distributions $f_x(\theta)$ and $g_x(\phi)$ for classifying sample $x$, as well as their ensemble $h_x(\underline{\theta},\phi)$, we have}
\begin{equation}
\begin{aligned}
\frac{\partial}{\partial \theta} D_{KL}(h_x(\underline{\theta},\phi)||f_x(\theta))=\beta \frac{\partial}{\partial \theta} D_{KL}(g_x(\phi)||f_x(\theta)).
\label{equ:proposition}
\end{aligned}
\end{equation} 
\textit{
Note that the variables with underscores do not participate in gradient propagation.}

The left-hand side of~\eqref{equ:proposition} suggests that when we update $\theta$ to make $f_x$ closer to $h_x$, we temporarily treat $h_x$ as a model with fixed parameters, though, in the next step, $h_x$ will also be updated with the new $\theta$.
According to the above proposition, we can replace the KL divergence term $D_{\text{KL}}(g_x(\phi)||f_x(\theta))$ in \eqref{equ:loss A} with $\frac{1}{\beta}D_{\text{KL}}(h_x(\underline{\theta},\phi)||f_x(\theta))$. Thus, we can represent \eqref{equ:loss A} as the sum of the cross-entropy loss and the KL divergence between $f_x(\theta)$ and $h_x(\underline{\theta},\phi)$:
\begin{equation}
\begin{aligned}
L_{f,x}(\theta)=L_{\text{CE}}(f_x(\theta))+\frac{\alpha}{\beta} D_{\text{KL}}(h_x(\underline{\theta},\phi)||f_x(\theta)).
\label{equ:modified loss}
\end{aligned}
\end{equation} 
From this perspective, we can see that KL divergence in MTE helps continuously align the primary model's predictions with the ensemble distribution during training, thereby improving the prediction diversity and producing well-calibrated uncertainty estimates.
By incorporating the principles of DE with the adaptive regularization provided by KL divergence, MTE offers a robust approach to improve the primary model's prediction accuracy and reliability. 

\section{Experiments}
We conducted primary experiments on image classification benchmarks, followed by experiments on misclassification and OOD detection to validate the performance of MTE across different tasks. 
Additionally, we performed experiments on noise-perturbed images and domain adaptation to further evaluate the robustness of MTE. 
\subsection{Experimental Setup}
\textbf{Datasets}:
We trained and evaluated our multi-class classification model using CIFAR-10, CIFAR-100,~\citet{krizhevsky2009learning} and Tiny-ImageNet~\citet{deng2009imagenet} datasets. Validation sets were created by randomly selecting 5,000 samples from each dataset's training set, with the remaining samples used for training. We experimented with two network architectures, ResNet34 and ResNet50~\citet{he2016deep}, on these datasets. To investigate MTE's calibration ability on noise-perturbed images, we used the CIFAR-10-C and CIFAR-100-C datasets~\citet{hendrycks2019benchmarking}, which introduce various types of corruptions at five different severity levels. Additionally, we tested the domain adaptation capability of our model using the PACS dataset~\citet{li2017deeper}, consisting of images from four domains: Art, Sketch, Cartoon, and Photo.

\textbf{Evaluation Metrics}:
For standard classification, we used accuracy (acc), Expected Calibration Error (ECE), and classwise-ECE (cw-ECE) to evaluate all methods. 
For misclassification and OOD detection, we used four metrics: (1) False positive rate at 95\% true positive rate (FPR-95\%); (2) The misclassification probability when true positive rate is 95\% (D-error); (3) Area under the receiver operating characteristic curve (AUROC); (4) Area under the precision-recall curve (AUPR).

\textbf{Competing Methods and Hyper-parameters}:
The evaluated methods include:
\textbf{Cross-Entropy (CE)}: This is the baseline method trained with CE loss~\eqref{equ:ce}. 
\textbf{MTE-1 \& MTE-2}: These refer to the MTE method using 1 and 2 auxiliary ResNet18 models, respectively. The primary model used was either ResNet34 or ResNet50. For both MTE-1 and MTE-2, we report the performance of the primary models.
The value $\alpha \in [0.4, 1.2]$ and we tuned it to achieve the best performance via a validation set.
\textbf{DCA}~\citet{liang2020improved}: Four values were used for its $\beta$: 0.5, 1.0, 5.0, and 10.0.
\textbf{MMCE}~\citet{kumar2018trainable}: Four values were used for its $\lambda$: 1.0, 2.0, 3.0, and 4.0.
\textbf{FL}~\citet{mukhoti2020calibrating}: Three values were used for its $\gamma$: 1.0, 2.0, and 3.0.
\textbf{FL+MDCA}~\citet{hebbalaguppe2022stitch}: 
Three values were used for the $\gamma$ in the focal loss: 1.0, 2.0, and 3.0, while the $\beta$ value in MDCA was 1.0.
\textbf{FL+DTDP}~\citet{patra2023calibrating}: Three values were used for the $\gamma$ in the focal loss: 1.0, 2.0, and 3.0. For DTDP, we used an $\alpha$ value of 0.005 and experimented with $\lambda$ values of 5.0 and 10.0, along with a data pruning ratio of 10\%.
\textbf{DE-3}~\citet{lakshminarayanan2017simple,fort2019deep}: It denotes DE with 3 models using cross-entropy loss with different initializations and combined their produced probabilities. 

\textbf{Implementation Details}: Please refer to Appendix~\ref{sec:details}.

\subsection{Classification Accuracy and Calibration}
\textbf{Comparison with Other Single-model Methods}:
Table~\ref{tab:cifar100} shows the comparison results of MTE and various single-model calibration methods on CIFAR-100. 
Tables~\ref{tab:cifar10} and~\ref{tab:tiny-imagenet} in Appendix~\ref{sec:table1} show the comparison results on CIFAR-10 and Tiny-ImageNet, respectively.
It can be observed that the majority of single-model methods compromised a certain level of accuracy to improve calibration ability. For instance, a previous state-of-the-art method, FL+MDCA, reduced accuracy by 1 to 2 percentage points. In contrast, our MTE method avoided this trade-off and demonstrates superior accuracy and calibration ability. 
For instance, on the CIFAR-100 dataset with ResNet34/50 models, compared to FL+MDCA, MTE-1 achieved absolute accuracy increases of $1.4\%/3.4\%$, relative reductions in ECE of $50.0\%/15.7\%$, and relative reductions in classwise-ECE of $7.7\%/10.8\%$. MTE-2 achieved absolute accuracy increases of $2.4\%/3.7\%$, relative reductions in ECE of $42.3\%/29.4\%$, and relative reductions in classwise-ECE of $11.6\%/15.3\%$, representing a considerable improvement.
\begin{table*}[!t]
\small
\caption{\small{Different calibration methods' accuracy, ECE, and cw-ECE on the CIFAR-100 dataset.} }
\begin{center}
\begin{tabular}{l|cc|cc|cc}
\toprule
\multirow{2}{*}{Method} &\multicolumn{2}{c|}{Acc(\%)$\uparrow$} &\multicolumn{2}{c|}{ECE(\%)$\downarrow$} &\multicolumn{2}{c}{cw-ECE($10^{-3}$)$\downarrow$}\\
&ResNet34 & ResNet50 & ResNet34 & ResNet50 & ResNet34 & ResNet50 \\
\midrule
Cross-Entropy & 76.93 & 77.54 & 8.81 & 10.39 & 2.25 & 2.51\\
DCA  & 73.63 & 72.64 & 6.53 & 5.85& 1.97 & 1.93 \\
MMCE & 74.50 & 74.40 & 6.58 & 4.71 & 1.94 & 1.72 \\
FL& 76.44 & 76.49 & 1.55 & 1.21 & 1.62& 1.57\\
FL+DTDP & 75.20 & 76.52 & 0.87  & 0.88 & 1.62 & 1.45 \\
FL+MDCA & 76.76 & 75.61 &1.30 & 1.02 & 1.55 & 1.57\\ 
\midrule
MTE-1(ours) & 78.20 & 79.05 & \textbf{0.65}  & 0.86 & 1.43 & 1.40 \\
MTE-2(ours) & \textbf{79.14} & \textbf{79.30} & 0.75  & \textbf{0.72} & \textbf{1.37} & \textbf{1.33} \\

\bottomrule
\end{tabular}
\end{center}
\label{tab:cifar100}
\end{table*}

\textbf{Comparison with DE}:
Table~\ref{tab:de} presents the accuracy and calibration results of the DE-3 method when integrating three ResNet34/50 models with different initializations. As a single-model inference approach, our MTE method achieved comparable accuracy while demonstrating superior confidence calibration and class-wise calibration abilities. Please note that MTE used one-third of the computational resources compared to DE-3 in inference, as the latter requires inference from multiple models. 
\label{sec:hist}
\begin{figure*}[t]
  \centering
  \includegraphics[width=\textwidth]{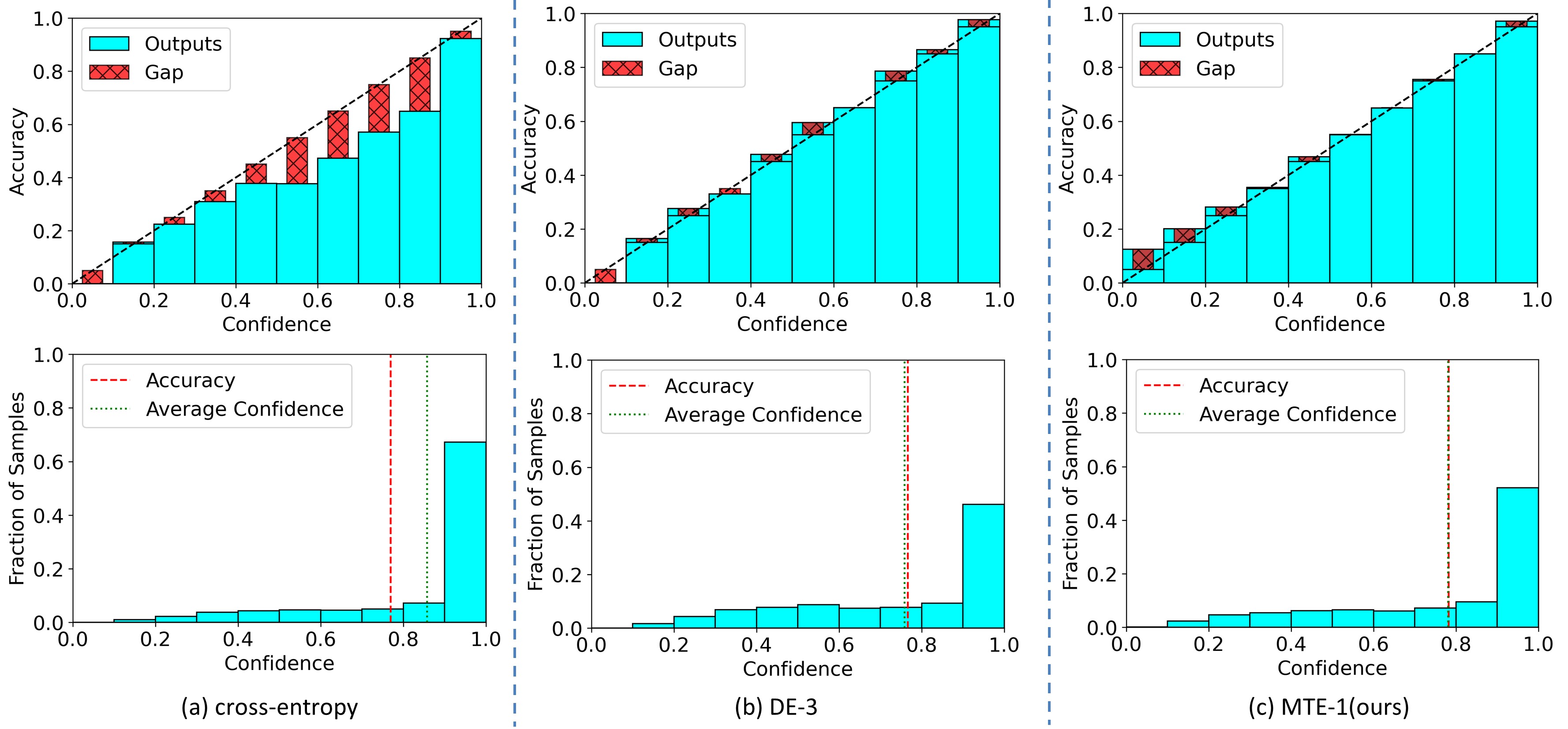}
  \caption{\small{Reliability diagrams (top) and confidence histograms (bottom) of (a) Cross Entropy, (b) DE-3, and (c) MTE-1 on CIFAR-100. In the reliability diagrams, blue bars depict the accuracy of model-predicted samples within various confidence intervals, while red bars signify the disparities between confidence and accuracy within the current probability interval. Ideally, a perfectly calibrated model would exhibit all blue bars aligned on the diagonal, implying the absence of red bars. Confidence histograms illustrate confidence distribution, with the green dashed line indicating average confidence and the red dashed line representing prediction accuracy.}}
\label{fig:conf-acc}
\end{figure*}
Figure~\ref{fig:conf-acc} displays the reliability diagrams and confidence histograms of CE, DE-3, and MTE-1 on the CIFAR-100 test set. In the reliability diagrams, the red bars represent the differences between confidence and accuracy within the current probability interval. It can be observed from the reliability diagrams that, compared to CE and DE-3, the length of most of the red bars in MTE-1 is shorter, indicating that the confidence predictions output by MTE-1 better aligned with its accuracy. From the confidence histograms, it is evident that, compared to cross-entropy, the number of samples with confidence higher than 90\% in the predicted distribution decreased noticeably with MTE training. Additionally, we present the average confidence and model accuracy for all samples in the confidence histograms, indicated respectively with the green dashed line and the red dashed line. It can be observed that the red and green dashed lines after MTE training are closer, indicating better calibration performance of MTE.

\begin{table*}[!t]
\small
\caption{\small{Comparison between DE and MTE on CIFAR-10, CIFAR-100, and Tiny-ImageNet datasets.} }
\begin{center}
\begin{tabular}{l|cc|cc|cc}
\toprule
\multirow{2}{*}{Method} &\multicolumn{2}{c|}{Acc(\%)$\uparrow$} &\multicolumn{2}{c|}{ECE(\%)$\downarrow$} &\multicolumn{2}{c}{cw-ECE($10^{-3}$)$\downarrow$}\\
&ResNet34 & ResNet50 & ResNet34 & ResNet50 & ResNet34 & ResNet50 \\
\midrule
\multicolumn{7}{l}{\textit{CIFAR-10}} \\
\midrule
DE-3& \textbf{95.79} & \textbf{95.63} & 1.19 & 0.77 & 3.64& 3.01\\
MTE-1(ours) & 95.39 & 95.20 & 0.68  & 0.90 & \textbf{2.43} & 2.60 \\
MTE-2(ours) & 95.29 & 95.23 & \textbf{0.51}  & \textbf{0.63} & 2.48 & \textbf{2.48} \\
\midrule
\multicolumn{7}{l}{\textit{CIFAR-100}} \\
\midrule
DE-3& \textbf{79.34} & \textbf{79.63} & 2.95 & 4.40 & 1.56 & 1.73\\
MTE-1(ours) & 78.20 & 79.05 & \textbf{0.65}  & 0.86 & 1.43 & 1.40 \\
MTE-2(ours) & 79.14 & 79.30 & 0.75  & \textbf{0.72} & \textbf{1.37}& \textbf{1.33} \\
\midrule
\multicolumn{7}{l}{\textit{Tiny-ImageNet}} \\
\midrule
DE-3& \textbf{58.78} & 57.60 & 1.95 & 3.32 & 1.58& 1.59\\
MTE-1(ours) & 57.50 & 58.02 & 1.33  & 1.20 & 1.52 & \textbf{1.48} \\
MTE-2(ours) & 58.60 & \textbf{58.08} & \textbf{1.08}  & \textbf{1.10} & \textbf{1.49} & 1.53 \\
\bottomrule
\end{tabular}
\end{center}
\label{tab:de}
\end{table*}

\subsection{Analysis of MTE}
\label{sec:analysis}

\textbf{Sensitivity of $\alpha$}: We examined the sensitivity of the hyper-parameter $\alpha$ in~\eqref{equ:loss A} and found that MTE is robust to variations in $\alpha$ (Figure~\ref{fig:alpha} in Appendix~\ref{sec:analysis details}). 
\textbf{Selection of Auxiliary Model}: We investigated how to select the auxiliary model and concluded that to ensure good performance, the auxiliary model's backbone should not significantly underperform compared to the primary model (Table~\ref{tab:model seletcion} in Appendix~\ref{sec:analysis details}).

\begin{table*}[!t]
\small
\caption{\small{Comparison of misclassification detection performance for different calibration methods. The optimal results are shown in bold, and the second-best results are underlined.} }
\begin{center}
\begin{tabular}{ll|cccc}
\toprule
Dataset & Method & FPR-95\%$\downarrow$ & D-error$\downarrow$& AUROC$\uparrow$&AUPR$\uparrow$\\ 
\midrule
\multirow{9}{*}{CIFAR-10} 
&Cross-Entropy & 38.53 & 13.20 & 91.80&99.46 \\
&DCA  &38.97 & 13.28 & 91.94 & 99.47 \\
&MMCE & 23.00 & 12.61 & 93.59 & 99.54\\
&FL &33.14 & 13.56 & 92.23 & 99.46\\
&FL+DTDP & 28.34 & 14.26 &92.20 & 99.36 \\
&FL+MDCA &32.80 & 13.21 & 92.41 & 99.47\\ 
&DE-3 & 22.37 & \underline{10.89} &94.08 & 99.68 \\
&MTE-1(ours) &\textbf{16.63} &\textbf{10.00} &\textbf{94.87} &\textbf{99.73}\\
&MTE-2(ours) &\underline{18.07}&11.14& \underline{94.46} &\underline{99.70} \\
\midrule
\multirow{9}{*}{CIFAR-100} 
&Cross-Entropy & 45.95 & 20.19 & 86.93 & 95.47\\
&DCA & 40.66 & 20.09 & 86.85 & 95.01 \\
&MMCE & 45.97& 20.02 & 87.00 &95.05 \\
&FL & 45.02 & 20.21 & 87.04 & 95.42 \\
&FL+DTDP & 43.95 &20.73 & 86.79 & 95.16 \\
&FL+MDCA &45.08 & 20.68 &86.56 & 95.36\\ 
&DE-3 & 40.04 & 19.25 &88.19 & 96.44 \\
&MTE-1(ours) &\underline{39.62} & \textbf{18.95} & \textbf{88.38} & \underline{96.50}\\
&MTE-2(ours) & \textbf{39.25} & \underline{19.12} & \underline{88.37} & \textbf{96.67}\\
\bottomrule
\end{tabular}
\end{center}
\label{tab:misclassification}
\end{table*}

\begin{table*}[!t]
\small
\caption{\small{Comparison of OOD detection performance for different calibration methods. The numbers before and after the backslash denote the near and far OOD detection results, respectively. The optimal results are shown in bold, and the second-best results are underlined.} }
\begin{center}
\begin{tabular}{ll|cccc}
\toprule
Dataset & Method & FPR-95\%$\downarrow$ & D-error$\downarrow$& AUROC$\uparrow$&AUPR$\uparrow$\\

\midrule
&Cross-Entropy & 69.48/19.80 & 19.45/10.59 & 86.11/94.60 & 82.00/72.54 \\
&DCA  & 64.63/18.38 & 18.80/10.10 & 87.07/95.05 & 84.36/72.48 \\
In-Distribution:&MMCE & 40.40/11.93 & 17.33/8.37 & 89.13/\textbf{97.53} &89.61/88.85 \\
CIFAR-10&FL & 66.13/19.20 & 20.97/10.69 & 85.57/94.28 & 82.81/74.18 \\
&FL+DTDP & 50.54/23.62 &20.38/13.54 & 86.57/92.03 & 86.73/77.10 \\
Out-of-Distribution:&FL+MDCA &72.19/29.84 & 20.75/13.58 &84.77/92.63 & 81.01/67.11\\ 
CIFAR-100/SVHN&DE-3 & 33.89/13.31 &15.60/9.07 & 90.12/95.06 & 90.27/86.53 \\
&MTE-1(ours) &\underline{30.12}/\textbf{10.51} & \textbf{15.06}/\textbf{7.65} & \textbf{91.87}/\underline{96.33} & \textbf{92.67}/\textbf{90.29}\\
&MTE-2(ours) & \textbf{29.87}/\underline{11.75} & \underline{15.14}/\underline{8.35} & \underline{91.75}/96.13 & \underline{92.62}/\underline{89.46}\\ 
\midrule
&Cross-Entropy & 68.26/45.87 & 28.69/22.27 & 77.30/85.44 & 77.49/54.01 \\
&DCA  &69.41/47.88 & 30.58/23.62 & 75.96/83.20 & 77.19/55.41 \\
In-Distribution:&MMCE & 65.34/49.38 & 29.53/25.77 & 76.58/77.64 & 78.66/49.53\\
CIFAR-100&FL &67.27/48.40 & 28.10/23.17 & 78.36/83.89 & 78.87/53.39 \\
&FL+DTDP & 59.10/53.85 & 27.23/24.16 &79.07/83.10 & 81.05/49.61 \\
Out-of-Distribution:&FL+MDCA &69.52/\underline{41.66} & 27.93/19.87 & 78.35/\textbf{87.46} & 77.92/59.08 \\ 
CIFAR-10/SVHN&DE-3 & \textbf{56.21}/48.70 & 26.61/24.20 &80.36/83.42 & 82.34/53.17 \\
&MTE-1(ours) & \textbf{56.21}/\textbf{37.84} &\textbf{25.55}/\textbf{19.58} &\textbf{81.50}/\underline{87.34} &\textbf{83.57}/\textbf{65.65}\\
&MTE-2(ours) &56.49/44.05&\underline{25.95}/\underline{20.72}& \underline{81.32}/86.93 &\underline{83.29}/\underline{61.38} \\

\bottomrule
\end{tabular}
\end{center}
\label{tab:ood}
\end{table*}

\subsection{Misclassification \& OOD Detection}
A critical real-world application of calibrated
predictions is to make the model aware of what it does not know. 
Specifically, the model should give low confidence for samples that are misclassified or deviate from the training distribution.  Motivated by this, we conducted experiments on misclassification detection and OOD detection. For these two tasks, we measured the success rate of detecting misclassified samples and OOD samples based on maximum softmax probability~\citet{hendrycks2016baseline}. We report the threshold-frees metrics (FPR-95\%, D-error, AUROC and AUPR).

\textbf{Misclassification Detection}: Misclassification detection is an important problem in machine learning, as it allows for the identification of instances where the model's predictions are unreliable. We expect the model to output low probability or high uncertainty for samples that cannot be classified correctly. Table~\ref{tab:misclassification} presents the performance results for various models in detecting misclassifications. Our method showed significant improvements over other single-model calibration techniques and the DE method.

\textbf{OOD Detection}:
A reliable classification model should exhibit higher prediction uncertainty and lower confidence when encountering test samples significantly different from the training data. We assessed different calibration methods' abilities to differentiate OOD samples by blending in-distribution test data with OOD data.
We assessed two capabilities of models trained on CIFAR-10 and CIFAR-100: far OOD detection and near OOD detection~\citet{fort2019deep,hendrycks2019scaling}. Far OOD detection involved distinguishing between CIFAR-10 and SVHN datasets~\citet{netzer2011reading} for models trained on CIFAR-10, and between CIFAR-100 and SVHN datasets for models trained on CIFAR-100. Near OOD detection required distinguishing between CIFAR-10 and CIFAR-100 datasets, which have similar domains.
The results, presented in Table~\ref{tab:ood}, demonstrate significant improvements of our method compared to other single-model calibration methods, even surpassing the performance of the DE method, known for its effectiveness in OOD detection.

\subsection{Calibration Performance on Noise-Perturbed Images}
\label{sec:cifar-c}
\begin{figure*}[!t]
  \centering
  \includegraphics[width=0.8\textwidth]{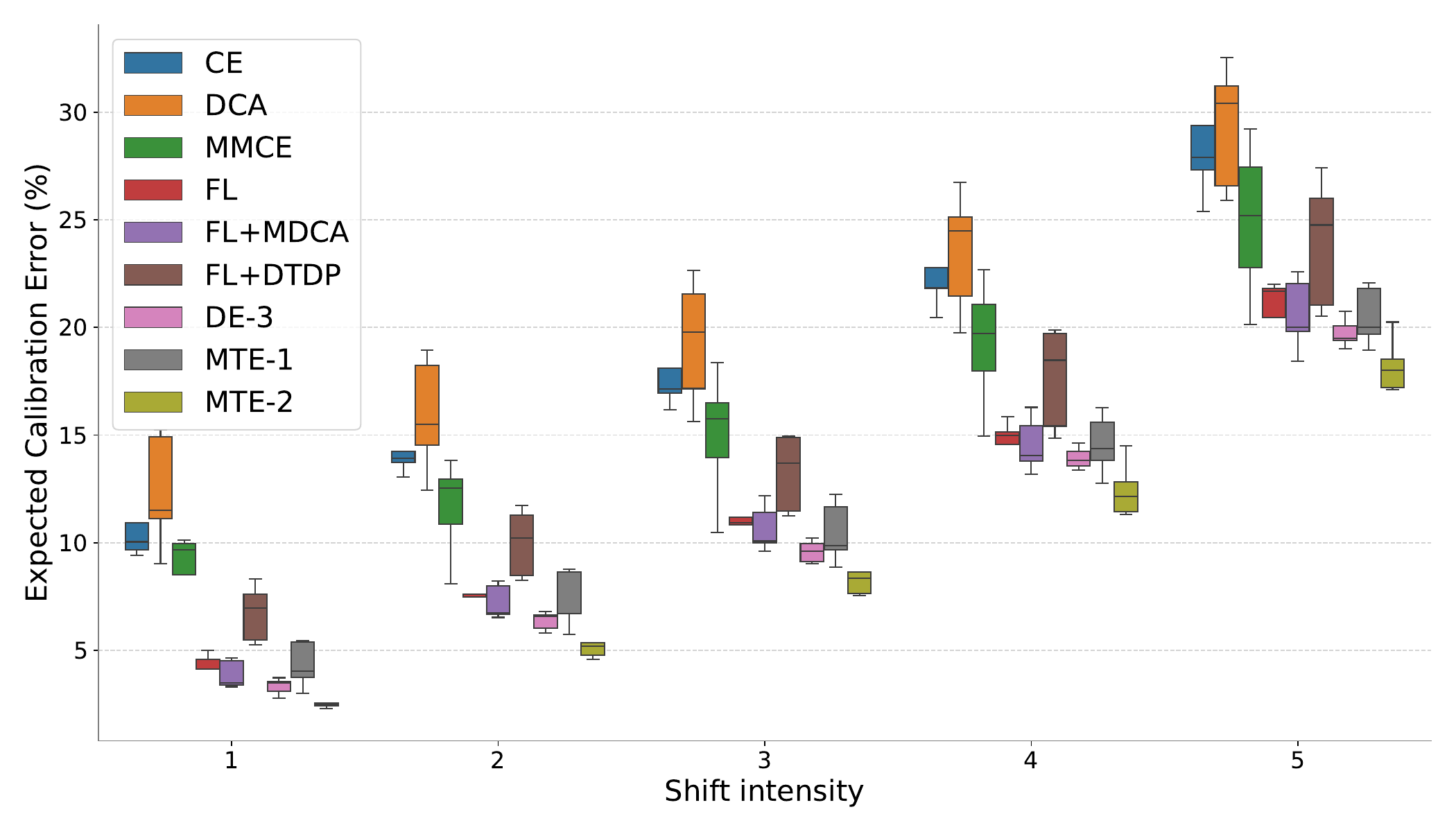}
  \caption{\small{Box plot: ECEs of different methods on
CIFAR-100-C under all types of corruptions with 5 levels of shift intensity. Each box shows a summary of the results of 16 types of shifts}}
\label{fig:cifar-c}
\end{figure*}

To investigate MTE's calibration performance on noise-perturbed images~\citet{ovadia2019can}, we evaluated the CIFAR-10/100 classifier using the CIFAR-10/100-C datasets, which introduce various types of corruptions at five different severity levels.
Figure~\ref{fig:cifar-c} illustrates ECE of each method across 16 different types of image corruptions, from level one (indicating the lowest severity) to level five (indicating the highest severity). It is concluded that MTE-1 achieved similar capabilities to state-of-the-art calibration methods when encountering distributional shift, while MTE-2 outperformed other methods in ECE across various degrees of shift. 

\subsection{Experiments on Domain Adaptation Capability}
\label{sec:domain adaptation}
Tomani et al.~\citet{tomani2021post} found that DNNs face decreased prediction accuracy and over-confident predictions under domain shift, leading to poor calibration. Therefore, effective calibration methods should adapt well to domain shifts. Our experiments showed that MTE outperformed other methods in domain adaptation for calibration. Details are provided in Appendix~\ref{sec:domain adaptation appendix}.

\subsection{Comparison with Deep Mutual Learning}
When comparing our MTE with Deep Mutual Learning (DML) proposed by Zhang et al.~\citet{zhang2018deep}, we found a key difference: DML trains both models with a combination of cross-entropy loss and KL divergence, while MTE trains the primary model with this combined loss and the auxiliary model with only KL divergence. Although MTE and DML had similar accuracy, MTE outperformed DML in calibration metrics.
Details are provided in Appendix~\ref{sec:dml details}.

\section{Limitation and Future Work}
\label{sec:limitation}
Although MTE effectively improves both accuracy and calibration, its reliance on an auxiliary model with performance comparable to the primary model poses challenges, particularly when training on models with large parameter sizes. This requirement can introduce difficulties in scalability and computational efficiency, especially in resource-constrained environments.
In future work, we aim to address this limitation by focusing on strategies to effectively implement MTE on models with large parameter sizes. By overcoming these challenges, we will extend the applicability of MTE to a wider range of deep neural network architectures, facilitating its adoption in real-world scenarios.

\section{Conclusion}
We introduce MTE, a novel training technique incorporating a co-trained auxiliary model and utilizing KL divergence as an adaptive regularizer. MTE markedly enhances both prediction accuracy and uncertainty calibration, crucial for ensuring reliable models in safety-critical contexts. Our experiments across diverse benchmarks consistently demonstrate MTE's efficacy. Unlike many existing approaches, MTE provides a scalable and efficient solution without compromising accuracy or imposing excessive computational burden, making it well-suited for real-world applications where reliability and efficiency are critical.

\bibliography{iclr2023_conference}
\bibliographystyle{iclr2023_conference}

\appendix
\appendix
\renewcommand{\thefigure}{S\arabic{figure}}
\renewcommand{\thetable}{S\arabic{table}}
\setcounter{figure}{0}  
\setcounter{table}{0}   

\section{Definition of ECE and Classwise-ECE}
\label{sec:background appendix}
In this section, we adopt the notation from Section~\ref{sec:method}.

ECE is defined as the expected absolute difference between the model’s confidence and its accuracy conditioned on confidence:
\begin{equation}
\begin{aligned}
{\text{ECE}}=\mathbb{E}_{x\sim P_{data}}( |\mathbb{E}_{x'\sim P_{data}}(\mathds{1}(\hat{y}_{x'}=y_{x'})|\hat{p}_{x'}=\hat{p}_{x})-\hat{p}_{x}| ),
\label{equ:ECE_inf}
\end{aligned}
\end{equation}
where $\mathbb{E}$ stands for expectation and $\mathds{1}$ is the indicator function.
Since we only have finite samples, ECE cannot be directly calculated using the definition provided above. Therefore, in practical calculations, we replace the above definition with a discretized version of ECE in which the interval $[0,1]$ is divided into $M$ equispaced bins. Let $B_i$ denote the samples with confidences belonging to the $i$-th bin (i.e. $(\frac{i-1}{M},\frac{i}{M}]$). The accuracy of this bin is $A_i=\frac{1}{|B_i|}\sum_{x\in B_i}\mathds{1}(\hat{y}_x=y_x)$. The average confidence of this bin is $C_i=\frac{1}{|B_i|}\sum_{x\in B_i}\hat{p}_x$. The discretized version of ECE is defined as
\begin{equation}
\begin{aligned}
{\text{ECE}}=\sum_{i=1}^M\frac{|B_i|}{N}|A_i-C_i|,
\label{equ:ECE_finite}
\end{aligned}
\end{equation}
where $N$ is the number of samples in the dataset.

Similar to ECE for confidence calibration, the classwise-ECE~\cite{kull2019beyond} for classwise calibration is defined as
\begin{equation}
\begin{aligned}
{\text{classwise-ECE}}=\frac{1}{K}\sum_{k=1}^K\mathbb{E}_{x\sim P_{data}}(|\mathbb{E}_{x'\sim P_{data}}(\mathds{1}(y_{x'}=k)|p_{x'}^{(k)}=p^{(k)}_x)-p^{(k)}_x|).
\label{equ:classwise ECE_inf}
\end{aligned}
\end{equation}
The discretized version of classwise-ECE is defined as
\begin{equation}
\begin{aligned}
{\text{classwise-ECE}}=\sum_{k=1}^K\sum_{i=1}^M\frac{|B_{i,k}|}{NK}|A_{i,k}-C_{i,k}|,
\label{equ:classwise ECE_inf}
\end{aligned}
\end{equation}
where $B_{i,k}$ denotes the set of samples whose predicted probabilities of the $k$-th class lie in the $i$-th bin,  $A_{i,k}=\frac{1}{|B_{i,k}|}\sum_{x\in B_{i,k}}\mathds{1}(y_x=k)$ and $C_{i,k}=\frac{1}{|B_{i,k}|}\sum_{x\in B_{i,k}}p^{(k)}_x$.

\section{Related Work}
\label{sec:related work}
\subsection{Single-Model Calibration Methods}
The current single-model calibration methods can be roughly divided into train-time calibration methods and post-hoc calibration methods.

\textbf{Train-time Calibration}: Train-time calibration methods focus on enhancing calibration during the training phase. These methods typically involve designing loss functions or incorporating additional regularization techniques.
Entropy Regularization (ER)~\cite{pereyra2017regularizing} is an approach that maximizes the entropy of the predicted probabilities while optimizing the cross-entropy loss. By encouraging the model to have more diverse and well-distributed predictions, it helps reduce overconfidence.
Label Smoothing (LS)~\cite{muller2019does,szegedy2016rethinking} optimizes the cross-entropy loss between the predicted probabilities and smoothed labels, instead of using hard labels. This encourages the model to learn a more nuanced understanding of the data distribution and improves calibration.
Focal Loss (FL)~\cite{mukhoti2020calibrating} addresses the issue of class imbalance by assigning more importance to minority class examples. By doing so, it improves calibration for both majority and minority classes.
Maximum Mean Calibration Error (MMCE)~\cite{kumar2018trainable} designs an auxiliary loss depended on the power of RKHS~\cite{gretton2013introduction} functions induced by a universal kernel.
Difference between Confidence and Accuracy (DCA)~\cite{liang2020improved} proposes an auxiliary loss that encourages the model to minimize the discrepancy between predicted confidence and accuracy.
Multi-class Difference in Confidence and Accuracy (MDCA)~\cite{hebbalaguppe2022stitch} extends the auxiliary loss introduced by DCA to calibrate the whole predicted probability distribution. This extension enhances the calibration performance of neural networks by ensuring accurate and reliable predictions across all confidence levels.
Dynamic Train-time Data Pruning (DTDP)~\cite{patra2023calibrating} achieves calibration by pruning low-confidence samples every few epochs. 

\textbf{Post-hoc Calibration}: Post-hoc calibration methods are applied after the model has been trained.
Temperature Scaling (TS)~\cite{platt1999probabilistic} is a commonly used method that smooths the logits of a deep neural network to achieve calibration. By adjusting the temperature parameter, TS aligns the predicted probabilities with the expected confidence of the model.
Dirichlet calibration (DC)~\cite{kull2019beyond} extends the Beta-calibration~\cite{kull2017beta} method from binary to multi-class classification. It models the predicted probabilities using a Dirichlet distribution and learns the parameters to calibrate the model.
Meta-calibration~\cite{bohdal2021meta} proposes a differentiable calibration method driven by ECE. 

\subsection{Deep Ensembles}
Deep Ensembles (DE)~\cite{lakshminarayanan2017simple,fort2019deep} is a widely used model ensemble technique in deep learning. DE creates an ensemble by training multiple models independently with different initializations or training data subsets. During inference, the predictions from all models are combined, typically by averaging their predicted probabilities.

DE has been shown to improve model performance and calibration by leveraging the diversity among independently trained models. Each model captures different aspects of the data and makes slightly different predictions, leading to more robust and well-calibrated ensemble predictions.

While DE has demonstrated effectiveness in improving model performance, it comes with additional computational overhead during inference.
In contrast to DE, Mutual-Transport Ensemble (MTE) incorporates a mutual feedback loop between the primary model and auxiliary models during training, leveraging the diversity among models while maintaining computational efficiency. This collaborative learning approach allows MTE to achieve improved calibration and performance without the need for maintaining multiple models during inference.

\section{Proof}
\label{sec:proof}
Here we provide the proof of the proposition 1 presented in the main text (\ref{equ:proposition}).

\textbf{Proposition 1}. 
\textit{Let $h_x(\underline{\theta},\phi)=(1-\beta)f_x(\underline{\theta})+\beta g_x(\phi)$ denote the ensemble predictive distribution of $f_x(\underline{\theta})$ and $g_x(\phi)$,$f_x(\underline{\theta})$ and $g_x(\phi)$. We have}
\begin{equation}
\begin{aligned}
\frac{\partial}{\partial \theta} D_{KL}(h_x(\underline{\theta},\phi)||f_x(\theta))=\beta \frac{\partial}{\partial \theta} D_{KL}(g_x(\phi)||f_x(\theta)),
\label{equ:proposition sm}
\end{aligned}
\end{equation} 
\textit{where $\beta \in [0,1]$ is the weight of the distributions.
Note that variables with underscores do not participate in gradient propagation.}

\textit{Proof}. Given a vector $v$, let $v^{(k)}$ denote the $k$-th dimension of $v$.
\begin{equation}
\begin{aligned}
\frac{\partial}{\partial \theta} D_{KL}(h_x(\underline{\theta},\phi)||f_x(\theta))&=\frac{\partial}{\partial \theta} \sum_{k=1}^K(h_x^{(k)}(\underline{\theta},\phi)\log \frac{h_x^{(k)}(\underline{\theta},\phi)}{f_x^{(k)}(\theta)}), \\
&=\sum_{k=1}^K \frac{\partial}{\partial \theta} [h_x^{(k)}(\underline{\theta},\phi)(\log h_x^{(k)}(\underline{\theta},\phi)-\log f_x^{(k)}(\theta))].
\end{aligned}
\end{equation} 
Since $\underline{\theta}$ does not participate in gradient backpropagation,
\begin{equation}
\begin{aligned}
\frac{\partial}{\partial \theta} D_{KL}(h_x(\underline{\theta},\phi)||f_x(\theta))&=-\sum_{k=1}^K h_x^{(k)}(\underline{\theta},\phi)\frac{\partial}{\partial \theta} \log f_x^{(k)}(\theta), \\
&=-(1-\beta)\sum_{k=1}^K f_x^{(k)}(\underline{\theta})\frac{\partial}{\partial \theta} \log f_x^{(k)}(\theta)-\beta \sum_{k=1}^K g_x^{(k)}(\phi)\frac{\partial}{\partial \theta} \log f_x^{(k)}(\theta).
\end{aligned}
\end{equation}
Let $z_x^{(i)}(\theta)$ denote the $i$-th dimension of the output logits of the primary model with input $x$. According to the chain rule of differentiation,
\begin{equation}
\begin{aligned}
\sum_{k=1}^K f_x^{(k)}(\underline{\theta})\frac{\partial}{\partial \theta} \log f_x^{(k)}(\theta)
&=\sum_{k=1}^K \frac{\partial}{\partial \theta} f_x^{(k)}(\theta) \\
&=\sum_{i=1}^K \sum_{k=1}^K \frac{\partial f_x^{(k)}(\theta)}{\partial z_x^{(i)}(\theta)} \frac{\partial z_x^{(i)}(\theta)}{\partial \theta}.
\end{aligned}
\end{equation}
Since $f_x^{(k)}(\theta)=\exp(z_x^{(k)}(\theta))/\sum_{j=1}^{K}\exp(z_x^{(j)}(\theta))$,
\begin{equation}
\begin{aligned}
\frac{\partial f_x^{(k)}(\theta)}{\partial z_x^{(i)}(\theta)}=\begin{cases}
f_x^{(i)}(\theta)(1-f_x^{(i)}(\theta)), & \text{where } k = i \\
-f_x^{(i)}(\theta)f_x^{(k)}(\theta), & \text{where } k \neq i
\end{cases}
\end{aligned}
\end{equation}
So,
\begin{equation}
\begin{aligned}
\sum_{k=1}^K \frac{\partial f_x^{(k)}(\theta)}{\partial z_x^{(i)}(\theta)} &=f_x^{(i)}(\theta)(1-f_x^{(i)}(\theta))-\sum_{k \neq i}f_x^{(i)}(\theta)f_x^{(k)}(\theta) \\
&=0.
\end{aligned}
\end{equation}
Such that
\begin{equation}
\begin{aligned}
\frac{\partial}{\partial \theta} D_{KL}(h_x(\underline{\theta},\phi)||f_x(\theta))
&=-\beta \sum_{k=1}^K g_x^{(k)}(\phi)\frac{\partial}{\partial \theta} \log f_x^{(k)}(\theta) \\
&=\beta \frac{\partial}{\partial \theta} \sum_{k=1}^K g_x^{(k)}(\phi)(\log g_x^{(k)}(\phi) - \log f_x^{(k)}(\theta)) \\
&=\beta \frac{\partial}{\partial \theta} D_{KL}(g_x(\phi)||f_x(\theta)).
\end{aligned}
\end{equation}
Q.E.D.
\section{Implementation Details} 
\label{sec:details}
In the experiments, we utilized the SGD optimizer with a momentum of 0.9 and weight decay of 0.0005. The initial learning rate was set to 0.1 for all datasets.
For each dataset, we trained the models for a total of 350 epochs. During the initial 100 epochs, we maintained the initial learning rate. Afterward, every 50 epochs, we decreased the learning rate by a factor of 10\%.
In the case of our MTE method, the primary model $f$ employed the same optimization strategy as described above. However, for the auxiliary model $g$, we initialized the learning rate as 0.01 and did not apply any learning rate decay.
All experiments conducted on CIFAR-10, CIFAR-100, and Tiny ImageNet were performed with a batch size of 100. 
However, due to memory constraints, a batch size of 20 was used for experiments on the PACS dataset. All experiments were conducted for training and testing on a single GeForce RTX 2080 GPU.

\section{Tables of Comparison Results on CIFAR-10 and Tiny-ImageNet}
\label{sec:table1}
See Tables~\ref{tab:cifar10} and~\ref{tab:tiny-imagenet}.
\begin{table*}[htbp]
\small
\caption{\small{Different calibration methods' accuracy, ECE, and cw-ECE on the CIFAR-10 dataset.} }
\begin{center}
\begin{tabular}{l|cc|cc|cc}
\toprule
\multirow{2}{*}{Method} &\multicolumn{2}{c|}{Acc(\%)$\uparrow$} &\multicolumn{2}{c|}{ECE(\%)$\downarrow$} &\multicolumn{2}{c}{cw-ECE($10^{-3}$)$\downarrow$}\\
&ResNet34 & ResNet50 & ResNet34 & ResNet50 & ResNet34 & ResNet50 \\
\midrule
Cross-Entropy   & 95.03 & 94.91 & 3.23 & 2.95 & 6.66 & 6.42  \\
DCA  & 95.22 & 94.91 & 3.28 & 3.22 & 7.01 & 6.68 \\
MMCE & 94.15 & 93.34 & 3.35 & 3.25 & 6.13 & 6.70\\
FL& 94.88 & 94.31 & 1.21 & 1.07 & 3.36& 4.06\\ 
FL+DTDP & 93.59 & 93.47& 1.41  & \textbf{0.59} & 3.99 & 3.31\\
FL+MDCA & 94.75 & 93.69 &0.87 &  0.68 & 3.04 & 3.72\\
\midrule
MTE-1(ours) & \textbf{95.39} & 95.20 & 0.68  & 0.90 & \textbf{2.43} & 2.60 \\
MTE-2(ours) & 95.29 & \textbf{95.23} & \textbf{0.51}  & 0.63 & 2.48 & \textbf{2.48} \\

\bottomrule
\end{tabular}
\end{center}
\label{tab:cifar10}
\end{table*}
\begin{table*}[htbp]
\small
\caption{\small{Different calibration methods' accuracy, ECE, and cw-ECE on the Tiny-ImageNet dataset.} }
\begin{center}
\begin{tabular}{l|cc|cc|cc}
\toprule
\multirow{2}{*}{Method} &\multicolumn{2}{c|}{Acc(\%)$\uparrow$} &\multicolumn{2}{c|}{ECE(\%)$\downarrow$} &\multicolumn{2}{c}{cw-ECE($10^{-3}$)$\downarrow$}\\
&ResNet34 & ResNet50 & ResNet34 & ResNet50 & ResNet34 & ResNet50 \\
\midrule
Cross-Entropy & 54.76 & 54.84 & 9.03 & 6.18 & 1.75 & 1.64\\
DCA  & 54.36 & 54.74 & 8.62 & 7.92 & 1.84 & 1.81 \\
MMCE & 54.50 & 54.12 & 6.55 & 4.37 & 1.67 & 1.56\\
FL& 53.28 & 53.38 & 1.19 & 2.13 & 1.61 & 1.63\\
FL+DTDP & 55.70 & 57.92 & 1.31  & 1.76 & 1.59 & 1.57 \\
FL+MDCA & 53.44 & 52.84 &1.33 & 2.16 & 1.67 & 1.67\\ 
\midrule
MTE-1(ours) & 57.50 & 58.02 & 1.33  & 1.20 & 1.52 & \textbf{1.48} \\
MTE-2(ours) & \textbf{58.60} & \textbf{58.08} & \textbf{1.08}  & \textbf{1.10} & \textbf{1.49} & 1.53 \\

\bottomrule
\end{tabular}
\end{center}
\label{tab:tiny-imagenet}
\end{table*}

\section{Details of Analysis}
\label{sec:analysis details}
\begin{figure*}[htbp]
  \centering
  \includegraphics[width=\textwidth]{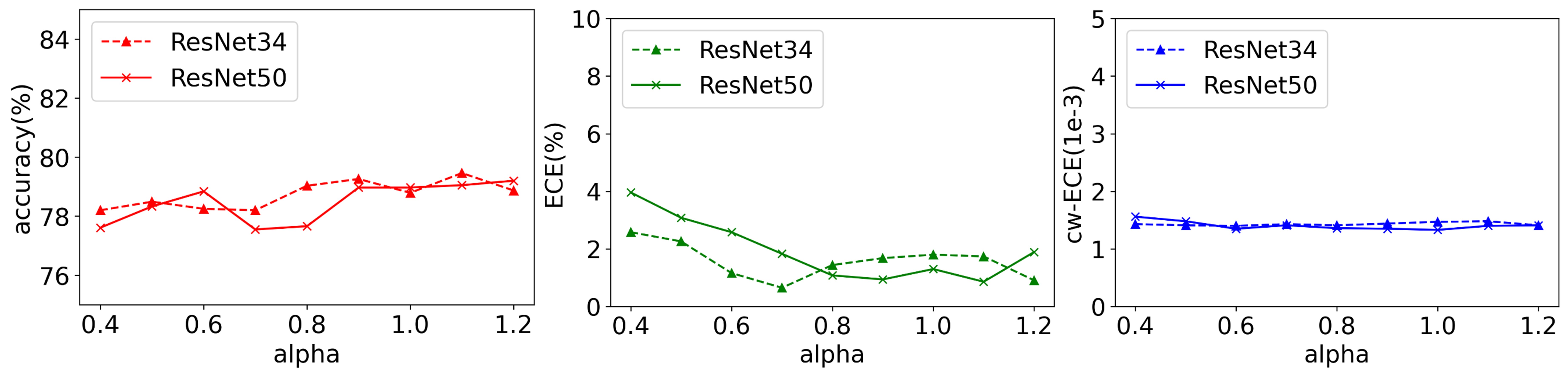}
  \caption{\small{Variation of accuracy, ECE, and classwise-ECE of MTE method trained on CIFAR-100 with different values of hyper-parameter $\alpha$}}
\label{fig:alpha}
\end{figure*}

\textbf{Sensitivity of $\alpha$}:
We investigated the sensitivity of the important hyper-parameter $\alpha$ in~\eqref{equ:loss A}.
From Figure~\ref{fig:alpha} it is observed that on CIFAR-100, as $\alpha$ changes, both accuracy and classwise-ECE exhibited smooth variations. Specifically, the fluctuations in accuracy remained within 3 percentage points, while the fluctuations in classwise-ECE were within 0.2 per mille. As for ECE, although its values ranged between 2 and 4 for very small (e.g., 0.4) and large (e.g., 1.2) $\alpha$ values, the fluctuations around the lowest point of ECE were also minimal. This demonstrates the good robustness of MTE with respect to the hyper-parameter $\alpha$.

\textbf{Selection of Auxiliary Model}:
We investigated how to select the auxiliary model. 
We experimented with CNNs containing two hidden layers, as well as ResNet18, ResNet34, and ResNet50 as backbones for the auxiliary model, which is shown in Table~\ref{tab:model seletcion}. It can be observed that only when using CNNs, both accuracy and calibration ability were significantly poor. When using ResNet18, ResNet34, and ResNet50, similar performance were observed. This indicates that in order to ensure good performance of the MTE, a backbone for the auxiliary model should not be selected if its performance is significantly worse than that of the primary model.

\begin{table*}[htbp]
\small
\caption{\small{Performance of MTE-1 using different backbones as auxiliary models} }
\begin{center}
\begin{tabular}{l|ccc|ccc}
\toprule
\multirow{2}{*}{\diagbox{$g$}{$f$}} &\multicolumn{3}{c|}{ResNet34} &\multicolumn{3}{c}{ResNet50} \\
&Acc(\%)$\uparrow$ & ECE(\%)$\downarrow$ & cw-ECE($10^{-3}$)$\downarrow$ & Acc(\%)$\uparrow$ & ECE(\%)$\downarrow$ & cw-ECE($10^{-3}$)$\downarrow$ \\
\midrule
\multicolumn{7}{l}{\textit{CIFAR-10}} \\
\midrule
CNN & 94.35 & 8.00 & 16.8 &94.10 & 7.65 & 16.14\\
ResNet18  & 95.39 & 0.68 & \textbf{2.34} & 95.20& 0.90 & 2.60 \\
ResNet34 & \textbf{95.48} & 0.65 & 2.35 & 95.16 & \textbf{0.59} & 2.59 \\
ResNet50& \textbf{95.48} & \textbf{0.57} & 2.56 & \textbf{95.25} & 0.64& \textbf{2.36}\\
\midrule
\multicolumn{7}{l}{\textit{CIFAR-100}} \\
\midrule
CNN & 76.09 & 5.31 & 2.03 & 75.36 & 7.67 & 2.42\\
ResNet18  & 78.20 & \textbf{0.65} & 1.43 & \textbf{79.05} & 0.86 & 1.40 \\
ResNet34 & \textbf{79.46} & 1.17 & \textbf{1.37} & 78.74 & 0.91 & \textbf{1.36} \\
ResNet50& 79.02 & 0.78 & 1.38 & 79.01 & \textbf{0.85}& 1.37\\
\bottomrule
\end{tabular}
\end{center}
\label{tab:model seletcion}
\end{table*}


\section{Details of Experiments on Domain Adaptation Capability}
\label{sec:domain adaptation appendix}
To validate MTE's domain adaptation capability in calibration, we conducted experiments on the PACS dataset, which serves as a benchmark for testing models' domain adaptation. Table~\ref{tab:pacs} presents the experimental results of different calibration methods on PACS, revealing that MTE outperformed other methods in terms of calibration's domain adaptation. It is worth mentioning that MTE was the only method among these calibration approaches that achieved superior accuracy and ECE metrics compared to cross-entropy baseline in all four domains.
\begin{table*}[htbp]
\small
\caption{\small{Accuracy and ECE of different calibration methods on the four domains in the PACS dataset.} }
\setlength{\tabcolsep}{3pt}
\begin{center}
\begin{tabular}{l|cc|cc|cc|cc}
\toprule
\multirow{2}{*}{Method} &\multicolumn{2}{c|}{Art} &\multicolumn{2}{c|}{Sketch} &\multicolumn{2}{c|}{Cartoon} &\multicolumn{2}{c}{Photo}\\
&Acc(\%)$\uparrow$ & ECE(\%)$\downarrow$ &Acc(\%)$\uparrow$ & ECE(\%)$\downarrow$&Acc(\%)$\uparrow$ & ECE(\%)$\downarrow$&Acc(\%)$\uparrow$ & ECE(\%)$\downarrow$ \\
\midrule
Cross-Entropy & 50.10 & 30.80 & 52.45 & 23.08 & 62.96  & 22.80 & 65.63 & 19.85\\
DCA  & 50.24 & 32.21 & 43.47  & 15.36 & 61.01 & 24.35 & 69.64 & 18.64 \\
MMCE & 44.53 & 13.86 & 57.65 & 8.50 & 59.47 & 17.68 & 68.80 &\textbf{4.70}\\
FL& 47.80 & \textbf{12.62} & 62.41 & 3.06 &60.88 & 14.86 & 69.34 & 9.90\\
FL+DTDP & 39.35 & 25.86 & 36.88  & 10.35 & 56.02 & 12.80 & 60.06 & 9.13 \\
FL+MDCA & 49.41 & 21.94 &55.99 & 4.55 & 58.61 & 18.80 & 69.70 & 11.15\\ 
\midrule
DE-3 & 52.69 & 19.58 & 44.59  & 8.35 & 65.91 & \textbf{6.91} & 71.98 & 8.39 \\
\midrule
MTE-1(ours) & \textbf{55.52} & 13.70 & 64.55  & 3.62 & 65.53 & 12.97 &\textbf{74.25} &6.00 \\
MTE-2(ours) & 53.27 & 13.51 & \textbf{70.58}  & \textbf{2.63} & \textbf{66.60} & 11.82 &73.77 &7.77 \\
\bottomrule
\end{tabular}
\end{center}
\label{tab:pacs}
\end{table*}

\section{Details of Comparison with DML}
\label{sec:dml details}
When comparing our MTE with Deep Mutual Learning (DML) proposed by Zhang et al.~\cite{zhang2018deep}, we found similarities with a key difference. In DML, both models were trained with a combination of cross-entropy loss and KL divergence, implying equal treatment during training. However, in our MTE, the primary model was trained with this combined loss, while the auxiliary model only used KL divergence as its loss.

As mentioned in Section~\ref{sec:intro}, cross-entropy loss can lead to miscalibration. By excluding this loss from the auxiliary model's training, we ensured it remained less overconfident, thus assisting the primary model in addressing overconfidence issues.

\begin{figure*}[t]
  \centering
  \includegraphics[width=\textwidth]{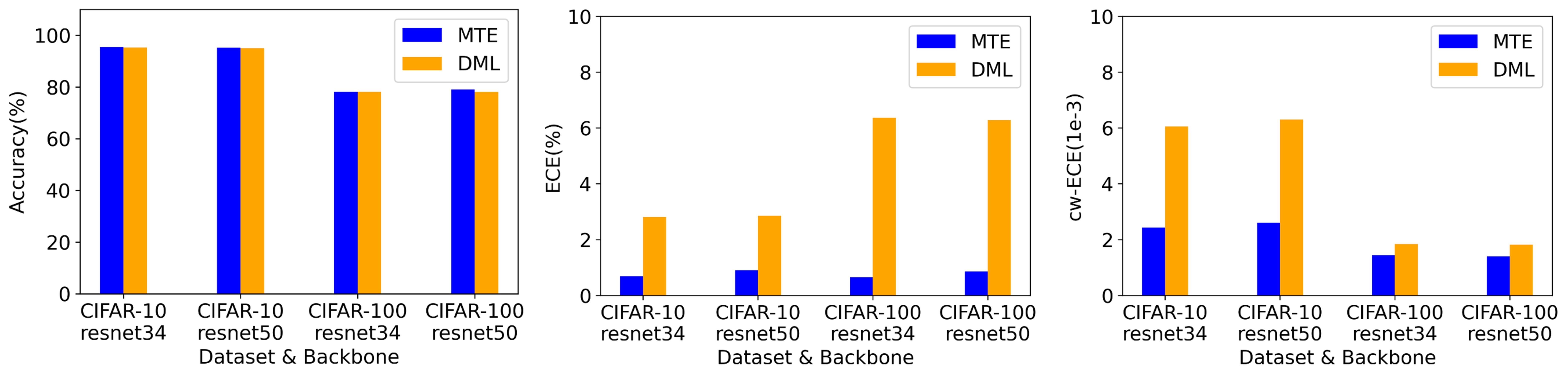}
  \caption{\small{The performance comparison of MTE and DML on accuracy, ECE, and CW-ECE metrics.}}
\label{fig:dml}
\end{figure*}

In Figure~\ref{fig:dml}, we present the performance of MTE and DML on the CIFAR-100 dataset. Since the original DML paper did not specify the hyper-parameter $\alpha$, we introduced it as the weight of the KL divergence loss in DML and determined the best $\alpha$ value based on validation set performance for a fair comparison.
From the figure, we observe that while MTE and DML showed similar accuracy, MTE outperformed DML in calibration metrics. This highlighted the effectiveness of our approach by leveraging the unique training strategy of the auxiliary model.

\end{document}

%% file: math_commands.tex
\usepackage{amsmath,amsfonts,bm}

\def\eqref#1{equation~\ref{#1}}

\def\1{\bm{1}}

\DeclareMathAlphabet{\mathsfit}{\encodingdefault}{\sfdefault}{m}{sl}
\SetMathAlphabet{\mathsfit}{bold}{\encodingdefault}{\sfdefault}{bx}{n}













%% file: iclr2023_conference.bbl
\begin{thebibliography}{33}
\providecommand{\natexlab}[1]{#1}
\providecommand{\url}[1]{\texttt{#1}}
\expandafter\ifx\csname urlstyle\endcsname\relax
  \providecommand{\doi}[1]{doi: #1}\else
  \providecommand{\doi}{doi: \begingroup \urlstyle{rm}\Url}\fi

\bibitem[Bohdal et~al.(2021)Bohdal, Yang, and Hospedales]{bohdal2021meta}
Ondrej Bohdal, Yongxin Yang, and Timothy Hospedales.
\newblock Meta-calibration: Meta-learning of model calibration using differentiable expected calibration error.
\newblock \emph{arXiv e-prints}, pp.\  arXiv--2106, 2021.

\bibitem[Cui et~al.(2024)Cui, Zhang, Deng, Dong, and Zhu]{cui2024learning}
Peng Cui, Dan Zhang, Zhijie Deng, Yinpeng Dong, and Jun Zhu.
\newblock Learning sample difficulty from pre-trained models for reliable prediction.
\newblock \emph{Advances in Neural Information Processing Systems}, 36, 2024.

\bibitem[Deng et~al.(2009)Deng, Dong, Socher, Li, Li, and Fei-Fei]{deng2009imagenet}
Jia Deng, Wei Dong, Richard Socher, Li-Jia Li, Kai Li, and Li~Fei-Fei.
\newblock Imagenet: A large-scale hierarchical image database.
\newblock In \emph{2009 IEEE conference on computer vision and pattern recognition}, pp.\  248--255. Ieee, 2009.

\bibitem[Dolezal et~al.(2022)Dolezal, Srisuwananukorn, Karpeyev, Ramesh, Kochanny, Cody, Mansfield, Rakshit, Bansal, Bois, et~al.]{dolezal2022uncertainty}
James~M Dolezal, Andrew Srisuwananukorn, Dmitry Karpeyev, Siddhi Ramesh, Sara Kochanny, Brittany Cody, Aaron~S Mansfield, Sagar Rakshit, Radhika Bansal, Melanie~C Bois, et~al.
\newblock Uncertainty-informed deep learning models enable high-confidence predictions for digital histopathology.
\newblock \emph{Nature communications}, 13\penalty0 (1):\penalty0 6572, 2022.

\bibitem[Fort et~al.(2019)Fort, Hu, and Lakshminarayanan]{fort2019deep}
Stanislav Fort, Huiyi Hu, and Balaji Lakshminarayanan.
\newblock Deep ensembles: A loss landscape perspective.
\newblock \emph{arXiv preprint arXiv:1912.02757}, 2019.

\bibitem[Greenman et~al.(2023)Greenman, Amini, and Yang]{greenman2023benchmarking}
Kevin~P Greenman, Ava~P Amini, and Kevin~K Yang.
\newblock Benchmarking uncertainty quantification for protein engineering.
\newblock \emph{bioRxiv}, pp.\  2023--04, 2023.

\bibitem[Gretton(2013)]{gretton2013introduction}
Arthur Gretton.
\newblock Introduction to rkhs, and some simple kernel algorithms.
\newblock \emph{Adv. Top. Mach. Learn. Lecture Conducted from University College London}, 16\penalty0 (5-3):\penalty0 2, 2013.

\bibitem[Guo et~al.(2017)Guo, Pleiss, Sun, and Weinberger]{guo2017calibration}
Chuan Guo, Geoff Pleiss, Yu~Sun, and Kilian~Q Weinberger.
\newblock On calibration of modern neural networks.
\newblock In \emph{International conference on machine learning}, pp.\  1321--1330. PMLR, 2017.

\bibitem[He et~al.(2016)He, Zhang, Ren, and Sun]{he2016deep}
Kaiming He, Xiangyu Zhang, Shaoqing Ren, and Jian Sun.
\newblock Deep residual learning for image recognition.
\newblock In \emph{Proceedings of the IEEE conference on computer vision and pattern recognition}, pp.\  770--778, 2016.

\bibitem[Hebbalaguppe et~al.(2022)Hebbalaguppe, Prakash, Madan, and Arora]{hebbalaguppe2022stitch}
Ramya Hebbalaguppe, Jatin Prakash, Neelabh Madan, and Chetan Arora.
\newblock A stitch in time saves nine: A train-time regularizing loss for improved neural network calibration.
\newblock In \emph{Proceedings of the IEEE/CVF Conference on Computer Vision and Pattern Recognition}, pp.\  16081--16090, 2022.

\bibitem[Hendrycks \& Dietterich(2019)Hendrycks and Dietterich]{hendrycks2019benchmarking}
Dan Hendrycks and Thomas Dietterich.
\newblock Benchmarking neural network robustness to common corruptions and perturbations.
\newblock \emph{arXiv preprint arXiv:1903.12261}, 2019.

\bibitem[Hendrycks \& Gimpel(2016)Hendrycks and Gimpel]{hendrycks2016baseline}
Dan Hendrycks and Kevin Gimpel.
\newblock A baseline for detecting misclassified and out-of-distribution examples in neural networks.
\newblock In \emph{International Conference on Learning Representations}, 2016.

\bibitem[Hendrycks et~al.(2019)Hendrycks, Basart, Mazeika, Zou, Kwon, Mostajabi, Steinhardt, and Song]{hendrycks2019scaling}
Dan Hendrycks, Steven Basart, Mantas Mazeika, Andy Zou, Joe Kwon, Mohammadreza Mostajabi, Jacob Steinhardt, and Dawn Song.
\newblock Scaling out-of-distribution detection for real-world settings.
\newblock \emph{arXiv preprint arXiv:1911.11132}, 2019.

\bibitem[Krizhevsky et~al.(2009)Krizhevsky, Hinton, et~al.]{krizhevsky2009learning}
Alex Krizhevsky, Geoffrey Hinton, et~al.
\newblock Learning multiple layers of features from tiny images.
\newblock 2009.

\bibitem[Kull et~al.(2017)Kull, Silva~Filho, and Flach]{kull2017beta}
Meelis Kull, Telmo Silva~Filho, and Peter Flach.
\newblock Beta calibration: a well-founded and easily implemented improvement on logistic calibration for binary classifiers.
\newblock In \emph{Artificial Intelligence and Statistics}, pp.\  623--631. PMLR, 2017.

\bibitem[Kull et~al.(2019)Kull, Perello~Nieto, K{\"a}ngsepp, Silva~Filho, Song, and Flach]{kull2019beyond}
Meelis Kull, Miquel Perello~Nieto, Markus K{\"a}ngsepp, Telmo Silva~Filho, Hao Song, and Peter Flach.
\newblock Beyond temperature scaling: Obtaining well-calibrated multi-class probabilities with dirichlet calibration.
\newblock \emph{Advances in neural information processing systems}, 32, 2019.

\bibitem[Kumar et~al.(2018)Kumar, Sarawagi, and Jain]{kumar2018trainable}
Aviral Kumar, Sunita Sarawagi, and Ujjwal Jain.
\newblock Trainable calibration measures for neural networks from kernel mean embeddings.
\newblock In \emph{International Conference on Machine Learning}, pp.\  2805--2814. PMLR, 2018.

\bibitem[Lakshminarayanan et~al.(2017)Lakshminarayanan, Pritzel, and Blundell]{lakshminarayanan2017simple}
Balaji Lakshminarayanan, Alexander Pritzel, and Charles Blundell.
\newblock Simple and scalable predictive uncertainty estimation using deep ensembles.
\newblock \emph{Advances in neural information processing systems}, 30, 2017.

\bibitem[Leibig et~al.(2017)Leibig, Allken, Ayhan, Berens, and Wahl]{leibig2017leveraging}
Christian Leibig, Vaneeda Allken, Murat~Se{\c{c}}kin Ayhan, Philipp Berens, and Siegfried Wahl.
\newblock Leveraging uncertainty information from deep neural networks for disease detection.
\newblock \emph{Scientific reports}, 7\penalty0 (1):\penalty0 1--14, 2017.

\bibitem[Li et~al.(2017)Li, Yang, Song, and Hospedales]{li2017deeper}
Da~Li, Yongxin Yang, Yi-Zhe Song, and Timothy~M Hospedales.
\newblock Deeper, broader and artier domain generalization.
\newblock In \emph{Proceedings of the IEEE international conference on computer vision}, pp.\  5542--5550, 2017.

\bibitem[Liang et~al.(2020)Liang, Zhang, Wang, and Jacobs]{liang2020improved}
Gongbo Liang, Yu~Zhang, Xiaoqin Wang, and Nathan Jacobs.
\newblock Improved trainable calibration method for neural networks on medical imaging classification.
\newblock \emph{arXiv preprint arXiv:2009.04057}, 2020.

\bibitem[Michelmore et~al.(2020)Michelmore, Wicker, Laurenti, Cardelli, Gal, and Kwiatkowska]{michelmore2020uncertainty}
Rhiannon Michelmore, Matthew Wicker, Luca Laurenti, Luca Cardelli, Yarin Gal, and Marta Kwiatkowska.
\newblock Uncertainty quantification with statistical guarantees in end-to-end autonomous driving control.
\newblock In \emph{International Conference on Robotics and Automation}, 2020.

\bibitem[Mukhoti et~al.(2020)Mukhoti, Kulharia, Sanyal, Golodetz, Torr, and Dokania]{mukhoti2020calibrating}
Jishnu Mukhoti, Viveka Kulharia, Amartya Sanyal, Stuart Golodetz, Philip Torr, and Puneet Dokania.
\newblock Calibrating deep neural networks using focal loss.
\newblock \emph{Advances in Neural Information Processing Systems}, 33:\penalty0 15288--15299, 2020.

\bibitem[M{\"u}ller et~al.(2019)M{\"u}ller, Kornblith, and Hinton]{muller2019does}
Rafael M{\"u}ller, Simon Kornblith, and Geoffrey~E Hinton.
\newblock When does label smoothing help?
\newblock \emph{Advances in neural information processing systems}, 32, 2019.

\bibitem[Naeini et~al.(2015)Naeini, Cooper, and Hauskrecht]{naeini2015obtaining}
Mahdi~Pakdaman Naeini, Gregory Cooper, and Milos Hauskrecht.
\newblock Obtaining well calibrated probabilities using bayesian binning.
\newblock In \emph{Proceedings of the AAAI conference on artificial intelligence}, volume~29, 2015.

\bibitem[Netzer et~al.(2011)Netzer, Wang, Coates, Bissacco, Wu, Ng, et~al.]{netzer2011reading}
Yuval Netzer, Tao Wang, Adam Coates, Alessandro Bissacco, Baolin Wu, Andrew~Y Ng, et~al.
\newblock Reading digits in natural images with unsupervised feature learning.
\newblock In \emph{NIPS workshop on deep learning and unsupervised feature learning}, volume 2011, pp.\ ~7. Granada, Spain, 2011.

\bibitem[Ovadia et~al.(2019)Ovadia, Fertig, Ren, Nado, Sculley, Nowozin, Dillon, Lakshminarayanan, and Snoek]{ovadia2019can}
Yaniv Ovadia, Emily Fertig, Jie Ren, Zachary Nado, David Sculley, Sebastian Nowozin, Joshua Dillon, Balaji Lakshminarayanan, and Jasper Snoek.
\newblock Can you trust your model's uncertainty? evaluating predictive uncertainty under dataset shift.
\newblock \emph{Advances in neural information processing systems}, 32, 2019.

\bibitem[Patra et~al.(2023)Patra, Hebbalaguppe, Dash, Shroff, and Vig]{patra2023calibrating}
Rishabh Patra, Ramya Hebbalaguppe, Tirtharaj Dash, Gautam Shroff, and Lovekesh Vig.
\newblock Calibrating deep neural networks using explicit regularisation and dynamic data pruning.
\newblock In \emph{Proceedings of the IEEE/CVF Winter Conference on Applications of Computer Vision}, pp.\  1541--1549, 2023.

\bibitem[Pereyra et~al.(2017)Pereyra, Tucker, Chorowski, Kaiser, and Hinton]{pereyra2017regularizing}
Gabriel Pereyra, George Tucker, Jan Chorowski, {\L}ukasz Kaiser, and Geoffrey Hinton.
\newblock Regularizing neural networks by penalizing confident output distributions.
\newblock \emph{arXiv preprint arXiv:1701.06548}, 2017.

\bibitem[Platt et~al.(1999)]{platt1999probabilistic}
John Platt et~al.
\newblock Probabilistic outputs for support vector machines and comparisons to regularized likelihood methods.
\newblock \emph{Advances in large margin classifiers}, 10\penalty0 (3):\penalty0 61--74, 1999.

\bibitem[Szegedy et~al.(2016)Szegedy, Vanhoucke, Ioffe, Shlens, and Wojna]{szegedy2016rethinking}
Christian Szegedy, Vincent Vanhoucke, Sergey Ioffe, Jon Shlens, and Zbigniew Wojna.
\newblock Rethinking the inception architecture for computer vision.
\newblock In \emph{Proceedings of the IEEE conference on computer vision and pattern recognition}, pp.\  2818--2826, 2016.

\bibitem[Tomani et~al.(2021)Tomani, Gruber, Erdem, Cremers, and Buettner]{tomani2021post}
Christian Tomani, Sebastian Gruber, Muhammed~Ebrar Erdem, Daniel Cremers, and Florian Buettner.
\newblock Post-hoc uncertainty calibration for domain drift scenarios.
\newblock In \emph{Proceedings of the IEEE/CVF Conference on Computer Vision and Pattern Recognition}, pp.\  10124--10132, 2021.

\bibitem[Zhang et~al.(2018)Zhang, Xiang, Hospedales, and Lu]{zhang2018deep}
Ying Zhang, Tao Xiang, Timothy~M Hospedales, and Huchuan Lu.
\newblock Deep mutual learning.
\newblock In \emph{Proceedings of the IEEE conference on computer vision and pattern recognition}, pp.\  4320--4328, 2018.

\end{thebibliography}
